\documentclass[11pt]{article}
\usepackage[margin=1in]{geometry}
\usepackage{times}
\usepackage{amsmath,amssymb}
\usepackage{booktabs}
\usepackage{multirow}
\usepackage{graphicx}
\usepackage{xcolor,colortbl}
\usepackage[numbers,sort&compress]{natbib}
\usepackage{microtype}
\usepackage{caption}
\usepackage{subcaption}
\usepackage{tikz}
\usetikzlibrary{arrows.meta,positioning,calc,shapes.geometric}
\usepackage{pgfplots}
\pgfplotsset{compat=1.17}
\usepgfplotslibrary{groupplots}
\usepackage[colorlinks=true,linkcolor=blue!50!black,citecolor=blue!50!black,urlcolor=blue!50!black]{hyperref}
\usepackage{fancyhdr}
\fancypagestyle{plain}{\fancyhf{}\fancyfoot[C]{\thepage}}
\definecolor{hl}{RGB}{232,243,232}
\definecolor{cfrozen}{RGB}{110,110,110}
\definecolor{cttt}{RGB}{30,120,60}
\definecolor{cfull}{RGB}{40,80,170}
\definecolor{chyb}{RGB}{190,80,30}
\definecolor{cdiff}{RGB}{130,40,150}
\newcommand{\best}[1]{\cellcolor{hl}#1}
\newcommand{\fw}{\mathbf{W}}

\pgfplotsset{every axis/.append style={font=\footnotesize, tick label style={font=\scriptsize}, label style={font=\footnotesize}, legend style={font=\scriptsize, draw=none, fill=none}}}

\begin{document}
\thispagestyle{plain}
\begin{center}
\rule{\textwidth}{1.2pt}\\[10pt]
{\LARGE\scshape Can Language Models Learn While They Think?\\[3pt] Testing Gradient Adaptation Across Architectures}\\[10pt]
\rule{\textwidth}{1.2pt}\\[18pt]
{\bfseries Surya Saka}\,$^{*1}$\\[2pt]
$^{1}$JudicialMind\\[22pt]
{\bfseries\scshape Abstract}
\end{center}
\vspace{-4pt}
\begin{list}{}{\leftmargin=1.6em \rightmargin=1.6em}\item\small
Letting a language model take gradient steps on the text it is reading, known as dynamic evaluation, lowers its loss on the rest of the document. The technique predates the current generation of open models, and its behaviour on hybrid linear-attention architectures, on diffusion language models, and on the benchmarks that long-context work is judged by has not been measured. We measure it. With one SGD step per 128-token chunk on the MLP output projections of the last quarter of blocks, no training, and a 2k-token window, softmax-attention models (Qwen3-1.7B, Qwen3-4B, Gemma-4-E4B) lose 5--8\% of their next-token loss over a document and outperform the same frozen model given the full 32k-token context. Hybrid Gated-DeltaNet models (Qwen3.5-4B, Qwen3.8-27B) gain under 1\% at every learning rate: the linear-attention blocks already perform the adaptation that the gradient step would add. The mechanism transfers to masked-diffusion language models with a larger effect, 8\% on Nemotron-Diffusion-8B and 17--20\% on DiffusionGemma-26B, once the rate is lowered and updates are clipped. Across RULER, BABILong, LongBench-v2, SCOTUS, HotpotQA, MuSiQue, ContractNLI and citation completion, at 100 or more items on two model sizes, weights updated on the full document recover none of the accuracy lost to context truncation: the adaptation is style and vocabulary, not retrieval. Writing facts with the next-token loss stores the training format rather than the fact (30\% recall under chat prompting against 80\% in context) and costs 36--38 GSM8K points at every dose. Writing a conclusion as behaviour instead, by distilling the model's own conclusion-conditioned judgments into a low-rank adapter, reaches the in-context ceiling (0.83 and 0.88 against 0.84 and 0.88 on 100 games, softmax and hybrid) with GSM8K unchanged, and the written conclusion can come from an inquiry loop that found it by asking information-gain questions. Every number is reproduced from per-item records in the released run logs.
\end{list}
\vspace{4pt}
\let\oldthefootnote\thefootnote
\renewcommand{\thefootnote}{}
\footnotetext{$^{*}$\texttt{Surya@judicialmind.ai}}
\let\thefootnote\oldthefootnote
\vspace{6pt}

\section{Introduction}

A language model that keeps learning while it reads is an old idea with a new audience. Dynamic evaluation~\citep{krause2019dynamic} adapts a model's parameters to the test stream by gradient descent. It was shown to lower perplexity on Transformer-XL. A 2024 revisit~\citep{rannentriki2024revisiting} confirmed the effect on 150M--1B models trained from scratch on PG-19. The revisit framed it as memory in weights versus memory in context. Since then the models people actually deploy have changed in three ways that bear directly on whether the idea still pays. Open models are now 2B--30B parameters and instruction-tuned. A growing share of them replace most softmax-attention blocks with gated linear attention~\citep{yang2024gated}. By construction it keeps a fast-weight state updated by a delta rule as tokens arrive~\citep{ttt_linear_attention}. Masked-diffusion language models have reached competitive quality without any autoregressive objective at all. None of these settings has a measurement of what inference-time gradient adaptation buys. The standard long-context benchmarks, on which a weight-based memory would have to show its value, have never been run against it.

We ran the technique on all three families and on the benchmarks with one fixed recipe and no training. The recipe was deliberately the minimum. We made the MLP output projection of the last quarter of transformer blocks writable. After each 128-token chunk was scored, we took one SGD step on the next-token loss. The attention window stayed at 2k tokens, and we reset the weights per document. Everything else in the model stayed frozen. This isolates the question of interest, what the gradient adds to what the architecture already does in context, from the separate question of what a model trained to use its fast weights could do~\citep{ttt_e2e,inplace_ttt}.

Four findings come out of 131 runs on two H100s. First, the adaptation gain is a property of the attention architecture. Softmax models lost 5--8\% of their loss over a book and beat the frozen model given the full 32k context, on every book tested. Hybrid Gated-DeltaNet models gained 0.3--0.8\% at the best rate and lost at higher rates. Second, the same gradient step worked on masked-diffusion language models. We ran it at a lower rate and clipped the updates. The effect was larger. We saw 8\% on a dense 8B diffusion model and 17--20\% on a 26B block-diffusion model. Third, the gain is not retrieval. The frozen model lost 6--50 points on eight benchmarks once we cut the context to the window. Updating the weights on the whole document first recovered none of them, at $n\ge100$ on a 4B and a 27B model. Fourth, the two routes by which one might store something specific in these weights diverged sharply. Next-token writes of facts memorised the prompt format and destroyed generated reasoning. A distillation write of a conclusion the model reached itself landed at the in-context ceiling with reasoning intact.

The contributions are the measurements and the controls that make them interpretable: a context-versus-adaptation control at 32k tokens on an off-the-shelf model, per-model learning-rate curves, a layer-location ablation, a surprise-gating ablation, a matched-drift comparison of two write objectives on the same parameters, and a composition in which an explicit Bayesian inquiry loop supplies the conclusion that gets written. Section~\ref{sec:related} places the work against dynamic evaluation, trained test-time-training layers, model editing and context distillation. Section~\ref{sec:method} fixes the mechanism and the evaluation protocol. Section~\ref{sec:exp} reports the results. Section~\ref{sec:discussion} gives the mechanism-level reading and the design rule that follows from it.

\section{Related Work}
\label{sec:related}

\paragraph{Dynamic evaluation and test-time training.}

\citet{krause2019dynamic} adapt all parameters of a Transformer-XL to the test sequence by gradient descent. They reduce WikiText-103 perplexity from 18.3 to 16.4. \citet{rannentriki2024revisiting} revisit the method on PG-19 with 150M--1B models. They update every 128 tokens with AdamW. Their 512-token context with adaptation matches a 2048-token static context. They also report that the gain concentrates under distribution shift. \citet{hardt2024ttt} fine-tune on retrieved neighbours of each test instance. We kept their framing and their 128-token update period. The measurement moved to current open models at 1.7B--27B parameters and to a 2k-versus-32k context control. We ran three attention architectures. Trained fast-weight memories are a different line. TTT layers~\citep{sun2024ttt} and Titans~\citep{behrouz2024titans} learn a memory module whose update rule is part of pretraining, while TTT-E2E~\citep{ttt_e2e} and In-Place TTT~\citep{inplace_ttt} train the whole model to use its inference-time updates and report retrieval gains after tens of billions of training tokens. Our measurements are the zero-training baseline those methods improve on.

\paragraph{Linear attention as fast weights.}

The Qwen3.5 and Qwen3.8 series use Gated DeltaNet~\citep{yang2024gated} and its hybrids with softmax attention. These blocks keep a matrix-valued state, updated by a gated delta rule. \citet{ttt_linear_attention} and MesaNet~\citep{mesanet} show formally that test-time training with key--value binding is a form of learned linear attention. This predicts that a gradient step on top of such blocks has little left to add. We provide the measurement the prediction lacked.

\paragraph{Diffusion language models at test time.}

We checked the test-time work on masked-diffusion LMs. It concerns sampling. Inference-time scaling~\citep{hex} is one line. Remasking schedules and reward guidance are others. Gradient adaptation of the denoiser on the document being processed has not been reported. We adapt the recipe to the denoising loss and to the context/canvas interface of block-diffusion models.

\paragraph{Model editing and its cost.}

ROME~\citep{meng2022rome} and MEMIT~\citep{meng2023memit} insert facts into mid-depth MLPs by closed-form rank-one updates. \citet{gu2024editing} show that editing under 1\% of parameters can collapse general ability, and sequential editing compounds the damage. Our next-token fact writes reproduce that cost with SGD and add the dose curve and the split between format collapse and reasoning loss.

\paragraph{Context distillation and self-generated updates.}

Context distillation~\citep{askell2021general,snell2022context} trains a model without a prompt to match its own outputs with the prompt. It is bounded by the prompted model. \citet{cd_memory2026} apply it per context into LoRA adapters as a memory system. SEAL~\citep{seal2025} has the model write its own fine-tuning data. Our behaviour write is context distillation into LoRA or late MLP projections. What we add is three things. We compare it with a next-token write on the same parameters at the same drift. We cover the hybrid models. And the distilled conclusion comes from an inquiry loop rather than a given document.

\section{Setting and Method}
\label{sec:method}

\subsection{Overview}

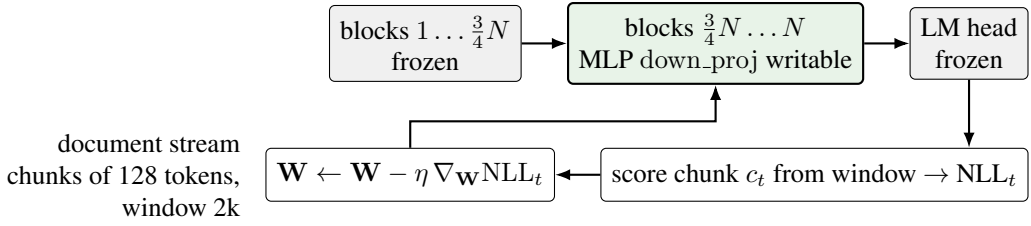
\begin{figure}[t]
\centering
\begin{tikzpicture}[font=\small, node distance=6mm,
  box/.style={draw, rounded corners=2pt, minimum height=7mm, inner sep=4pt, align=center},
  frozen/.style={box, fill=gray!12},
  fast/.style={box, fill=hl, thick},
  arr/.style={-{Latex[length=2mm]}, thick}]
\node[frozen] (emb) {blocks $1 \ldots \tfrac{3}{4}N$\\ frozen};
\node[fast, right=of emb] (late) {blocks $\tfrac{3}{4}N \ldots N$\\ MLP $\mathrm{down\_proj}$ writable};
\node[frozen, right=of late] (head) {LM head\\ frozen};
\node[box, anchor=north east, fill=white] (loss) at ([yshift=-9mm]head.south east) {score chunk $c_t$ from window $\rightarrow$ NLL$_t$};
\node[box, left=of loss, fill=white] (step) {$\fw \leftarrow \fw - \eta\,\nabla_{\fw}\mathrm{NLL}_t$};
\draw[arr] (emb) -- (late);
\draw[arr] (late) -- (head);
\draw[arr] (head.south) -- (head.south |- loss.north);
\draw[arr] (loss) -- (step);
\draw[arr] (step.north) -- ++(0,3.5mm) -| (late.south);
\node[left=2mm of step, align=right] {document stream\\ chunks of 128 tokens,\\ window 2k};
\end{tikzpicture}
\caption{Inference-time fast weights. Each chunk is scored from a 2k window before the single SGD step that uses its loss; only the late MLP output projections move; $\fw$ is reset per document.}
\label{fig:overview}
\end{figure}

\begin{figure}[t]
\centering
\begin{tikzpicture}[font=\small, node distance=4mm and 8mm,
  doc/.style={draw, fill=gray!8, minimum height=6mm, minimum width=38mm, inner sep=2pt},
  win/.style={draw, fill=hl, minimum height=6mm, minimum width=10mm, inner sep=2pt},
  model/.style={draw, rounded corners=2pt, fill=white, minimum height=7mm, inner sep=4pt, align=center},
  arr/.style={-{Latex[length=2mm]}, thick}]
\node[doc] (d1) {whole document (up to 64k)};
\node[model, right=of d1] (m1) {frozen $f_\theta$};
\node[right=of m1] (a1) {answer};
\node[left=2mm of d1, align=right, font=\footnotesize\itshape] {Full};
\draw[arr] (d1) -- (m1); \draw[arr] (m1) -- (a1);
\node[doc, below=of d1, fill=white, draw=gray!50, dashed] (d2) {};
\node[win, anchor=east] (w2) at (d2.east) {last 2k};
\node[model, right=of d2] (m2) {frozen $f_\theta$};
\node[right=of m2] (a2) {answer};
\node[left=2mm of d2, align=right, font=\footnotesize\itshape] {Window};
\draw[arr] (w2.east) -- (m2); \draw[arr] (m2) -- (a2);
\node[doc, below=of d2] (d3) {whole document, read in chunks};
\node[model, right=of d3, fill=hl] (m3) {$f_\theta$ with $\fw$ updating};
\node[doc, below=of d3, fill=white, draw=gray!50, dashed] (d4) {};
\node[win, anchor=east] (w3) at (d4.east) {last 2k};
\node[model, right=of d4, fill=hl] (m4) {$f_\theta$ with updated $\fw$};
\node[right=of m4] (a3) {answer};
\node[left=2mm of d3, align=right, font=\footnotesize\itshape] {Fast weights};
\draw[arr] (d3) -- (m3); \draw[arr] (m3) -- (m4); \draw[arr] (w3.east) -- (m4); \draw[arr] (m4) -- (a3);
\end{tikzpicture}
\caption{The three-condition protocol. Full $-$ Window is what the document holds beyond the window; FW $-$ Window is what the updated weights recovered of it. All three conditions score the same items.}
\label{fig:protocol}
\end{figure}
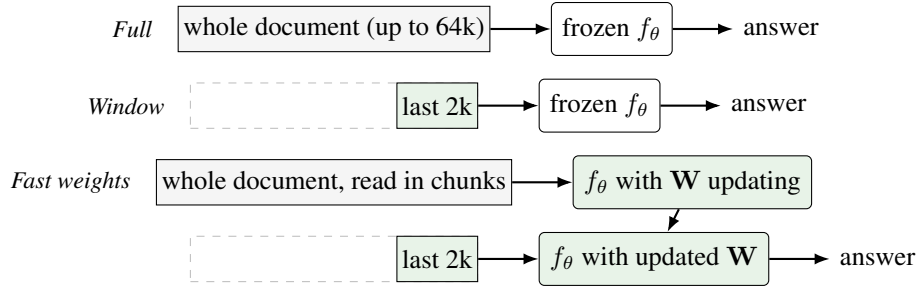

Given a frozen pretrained language model $f_\theta$ and a document $x_{1:T}$ that exceeds the attention window $L$, the task is to predict each chunk of the document from the preceding window alone, and to answer questions about the document from the window alone. We add a small writable parameter set $\fw$ that is updated by gradient descent as the document is read (Section~\ref{sec:fastweights}), fix a three-condition protocol that separates what the window gives, what the full context gives, and what the updated weights give (Section~\ref{sec:protocol}), define two ways to write a specific conclusion into $\fw$ (Section~\ref{sec:writes}), and specify the inquiry loop that produces the conclusion (Section~\ref{sec:inquiry}). Figure~\ref{fig:overview} shows the reading mechanism and Figure~\ref{fig:protocol} the protocol.

\subsection{Inference-time fast weights}
\label{sec:fastweights}

The writable set is the MLP output projection of the last quarter of transformer blocks. For a model with $N$ blocks this is $\{\mathrm{down\_proj}_\ell : \ell \ge \lfloor 3N/4 \rfloor\}$; for Qwen3-8B it is 453M parameters, for Nemotron-Diffusion-8B 470M, for DiffusionGemma-26B-A4B the last 8 of 30 encoder blocks, 48M. The choice is fixed before any experiment and ablated in Section~\ref{sec:controls}: the first quarter gives nothing, the middle quarter nearly matches the last.

We read the document in chunks $c_1, c_2, \ldots$ of 128 tokens. 256 tokens went to legal text and diffusion models. Benchmark documents used 512 tokens. For chunk $c_t$ the model computed the next-token loss $\mathrm{NLL}_t$ with the preceding $L=2048$ tokens as context and the current $\fw$. We recorded that loss as the score for $c_t$. Only after that did we take one SGD step, $\fw \leftarrow \fw - \eta\,\nabla_{\fw}\mathrm{NLL}_t$. Scoring before updating made every reported loss a genuine prediction of unseen text. Per chunk we recomputed the window without a key--value cache, so the attention context stayed exactly $L$ tokens throughout. We kept a master copy of $\fw$ in fp32 for bf16 models. Updates of relative size $10^{-4}$ were otherwise lost to rounding. At the start of each document we reset $\fw$ to its pretrained value and logged the drift $\|\fw - \fw_0\|/\|\fw_0\|$. There was no momentum and no forgetting term. We left the learning-rate schedule out. A surprise-dependent gate and multiple inner steps were evaluated as ablations. Both were rejected.

For masked-diffusion models we used the model's own training objective as the loss. We masked each chunk at a random ratio $t \sim U(0,1)$. The denoiser predicted the masked tokens from the unmasked ones plus the window. The loss was the $1/t$-weighted masked cross-entropy. For block-diffusion models with a separate context and canvas interface, the window entered as context and the chunk as the masked canvas. The denoising loss gave larger gradients, so we clipped per-step updates to a relative norm of 0.5.

\subsection{Three-condition protocol}
\label{sec:protocol}

We report every benchmark result as a triple on the same items. Under \emph{Frozen, full context}, the frozen model saw the whole document, up to 64k tokens, and answered. For \emph{Frozen, window}, we clipped the document to its last $L$ tokens before the frozen model answered. In \emph{Fast weights, window}, we ran the model on the whole document in chunks with $\fw$ updating. It then answered from the last $L$ tokens with the updated $\fw$. The gap between the first two conditions is what the document contains that the window does not; the gap between the last two is what the weights recovered of it. For perplexity we ran the same three conditions, with per-chunk loss as the metric. The full-context condition was a single forward pass over the document at the model's native window.

Multiple-choice tasks are scored by raw completion log-likelihood of each option after the prompt, with no chat template. Generation tasks use greedy decoding and exact match, SQuAD-normalised F1, or the RULER criterion that every expected answer string must appear. Every per-item prediction is logged and every reported mean is recomputed from those records.

\subsection{Writing a specific conclusion}
\label{sec:writes}

\begin{figure}[t]
\centering
\begin{tikzpicture}[font=\small, node distance=5mm and 7mm,
  box/.style={draw, rounded corners=2pt, minimum height=7mm, inner sep=4pt, align=center, fill=white},
  bank/.style={box, fill=gray!10},
  write/.style={box, fill=hl, thick},
  probe/.style={box, dashed},
  arr/.style={-{Latex[length=2mm]}, thick}]
\node[bank] (bank) {hypothesis bank\\ posterior $p(h)$};
\node[box, right=of bank] (q) {pick question\\ max EIG};
\node[box, right=of q] (env) {environment\\ answers};
\node[box, right=of env] (lik) {judge likelihood\\ $p(\text{ans}\mid h)$, few-shot};
\draw[arr] (bank) -- (q); \draw[arr] (q) -- (env); \draw[arr] (env) -- (lik);
\draw[arr] (lik.south) -- ++(0,-5mm) -| (bank.south) node[pos=0.25, below, font=\footnotesize] {Bayes update; stop at $\max_h p(h) \ge 0.9$};
\node[box, below=14mm of q, fill=gray!10] (concl) {conclusion $s = \arg\max_h p(h)$};
\draw[arr] (bank.south) ++(-4mm,0) |- (concl.west);
\node[write, below left=8mm and -10mm of concl] (tw) {text write\\ SGD on tokens of $s$};
\node[write, below right=8mm and -10mm of concl] (bw) {behaviour write\\ $s$-conditioned labels $\rightarrow$ KL into $\fw$/LoRA};
\draw[arr] (concl) -- (tw); \draw[arr] (concl) -- (bw);
\node[probe, below=6mm of tw] (p1) {new cases, no $s$ in prompt\\ AUROC vs true rule};
\node[probe, below=6mm of bw] (p2) {collateral: other-rule prompt-following,\\ GSM8K / MMLU / IFEval};
\draw[arr] (tw) -- (p1); \draw[arr] (bw) -- (p2);
\draw[arr] (tw.south east) -- (p2.north west); \draw[arr] (bw.south west) -- (p1.north east);
\end{tikzpicture}
\caption{The composition. An inquiry loop finds a conclusion by information-gain questioning; the conclusion is written into the weights by one of two objectives; both writes face the same behavioural and collateral probes.}
\label{fig:composition}
\end{figure}
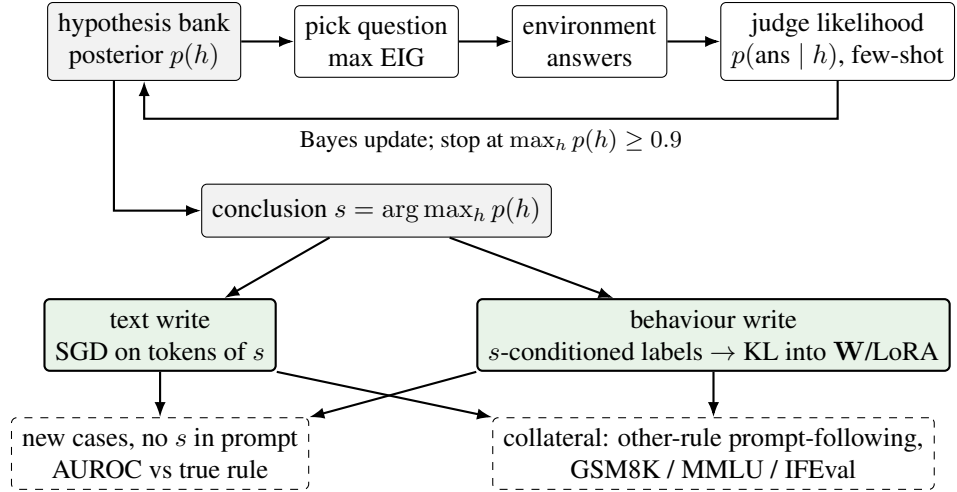

We had two objectives that could put a specific statement $s$ into $\fw$. In the \emph{text write}, we ran the fast-weight step on the tokens of $s$ itself. We repeated it for $p$ passes at rate $\eta$. That is the mechanism of Section~\ref{sec:fastweights} applied to a sentence. For the \emph{behaviour write}, we ran context distillation. With $s$ in its prompt, the model labeled a set of fresh inputs. We trained $\fw$, or a rank-16 LoRA over all linear layers, with Adam on the soft KL between the model's label distribution without $s$ and the recorded labels. Both writes moved the same parameters by a comparable relative amount. They differed only in objective. After each write, we ran a collateral probe. We checked the model's accuracy on a held-out instruction unrelated to $s$, and its GSM8K, MMLU and IFEval scores. Figure~\ref{fig:composition} places both writes in the full loop.

\subsection{The inquiry loop}
\label{sec:inquiry}

The conclusion $s$ comes from an inquiry loop over an explicit hypothesis bank. The bank holds $K$ candidate hypotheses with a posterior $p(h)$, uniform at the start. Each observation $(q, a)$ updates the posterior by Bayes' rule with the model's own likelihood $p(a \mid q, h)$, obtained by teacher-forcing the yes/no answer under a prompt that states $h$. A four-shot prefix with rules outside the bank is part of that prompt; without it the models answer ``yes'' 95\% of the time regardless of $h$, and the posterior is uninformative. The next question is chosen from a candidate set by expected information gain, $\mathrm{EIG}(q) = H[p(h)] - \mathbb{E}_{a}\,H[p(h \mid q, a)]$, with the expectation under the current predictive distribution. The loop stops when the maximum posterior exceeds 0.9 or after 30 questions, and emits $s = \arg\max_h p(h)$. In the environment used here the hypotheses are 24 enumerable rules over integer triples, the question candidates are objects from a space of 729, and an exact oracle likelihood is available as a reference for the judge.

\section{Experiments}
\label{sec:exp}

\subsection{Setup}

\textbf{Models.} For softmax attention we ran Qwen3-1.7B, Qwen3-4B, and Qwen3-8B~\citep{qwen3}, plus Gemma-4-E4B~\citep{gemma4}. The hybrid Gated-DeltaNet uses three of every four blocks as linear attention; we ran Qwen3.5-4B, Qwen3.5-9B, and Qwen3.8-27B~\citep{qwen38}. For masked diffusion we ran Nemotron-Labs-Diffusion-8B~\citep{nemotron_diffusion}, dense with block size 32, and DiffusionGemma-26B-A4B-it~\citep{diffusiongemma}, block diffusion with a context/canvas interface. Qwen3-1.7B ran in fp32. We kept everything else in bf16, with fp32 master copies of $\fw$.

\textbf{Data.} Adaptation is measured on the PG-19 test split~\citep{rae2019pg19}, 32k tokens per book (16k for diffusion models), and on a stream of U.S.\ court opinions of at least 30k characters from the Pile of Law CourtListener shard~\citep{henderson2022pileoflaw}, de-duplicated. Benchmarks: RULER~\citep{hsieh2024ruler} at 4k and 16k, BABILong~\citep{kuratov2024babilong} 16k qa1, LongBench-v2~\citep{bai2024longbenchv2}, SCOTUS and CaseHOLD from LexGLUE~\citep{chalkidis2022lexglue,zheng2021casehold}, LegalBench hearsay and consumer-contracts QA~\citep{guha2023legalbench}, ContractNLI~\citep{koreeda2021contractnli}, HotpotQA~\citep{yang2018hotpotqa}, MuSiQue~\citep{trivedi2022musique}, CounterFact~\citep{meng2022rome}, zsRE from KnowEdit~\citep{zhang2024knowedit}, and a citation-completion task built from the court opinions (the model must produce the next reporter citation given the opinion so far). Retention uses GSM8K, MMLU and IFEval through lm-eval.

\textbf{Compute.} All runs use two H100 NVL 96GB GPUs, about 19 GPU-hours in total. The adaptation pass costs 1.34--1.49$\times$ the frozen pass in wall-clock.

\subsection{Adaptation across architectures}
\label{sec:adapt}

\begin{table}[t]
\caption{Inference-time adaptation on PG-19 and on court opinions. Gain is the mean over documents of (frozen NLL $-$ fast-weight NLL) in nats per token with a 95\% CI, at each model's best learning rate; ``improved'' counts documents whose loss fell. Window 2k, chunk 128 (256 for opinions), one SGD step per chunk, no training.}
\label{tab:adapt}
\centering\small
\begin{tabular}{llrrrcrr}
\toprule
Model & Attention & $n$ docs & $\eta$ & Gain (nats) $\uparrow$ & Rel. & Improved & Drift \\
\midrule
\multicolumn{8}{l}{\textit{PG-19 books, 32k tokens each}}\\
Qwen3-1.7B & softmax & 98 & $10^{-2}$ & \best{$+0.216 \pm 0.007$} & 7\% & 98/98 & 0.19\% \\
Qwen3-4B & softmax & 20 & $10^{-2}$ & \best{$+0.227 \pm 0.017$} & 8\% & 20/20 & 0.20\% \\
Gemma-4-E4B & softmax & 20 & $10^{-3}$ & \best{$+0.145 \pm 0.020$} & 5\% & 20/20 & 0.17\% \\
Qwen3.5-4B & hybrid GDN & 20 & $10^{-3}$ & $+0.012 \pm 0.002$ & 0.5\% & 20/20 & 0.12\% \\
Qwen3.8-27B & hybrid GDN & 10 & $10^{-2}$ & $+0.019 \pm 0.006$ & 0.8\% & 10/10 & 0.21\% \\
\midrule
\multicolumn{8}{l}{\textit{Court opinions, $\ge$30k characters}}\\
Qwen3-1.7B & softmax & 100 & $10^{-2}$ & \best{$+0.152 \pm 0.004$} & 7\% & 100/100 & 0.11\% \\
Qwen3-4B & softmax & 100 & $10^{-2}$ & \best{$+0.143 \pm 0.003$} & 7\% & 100/100 & 0.12\% \\
\bottomrule
\end{tabular}
\end{table}

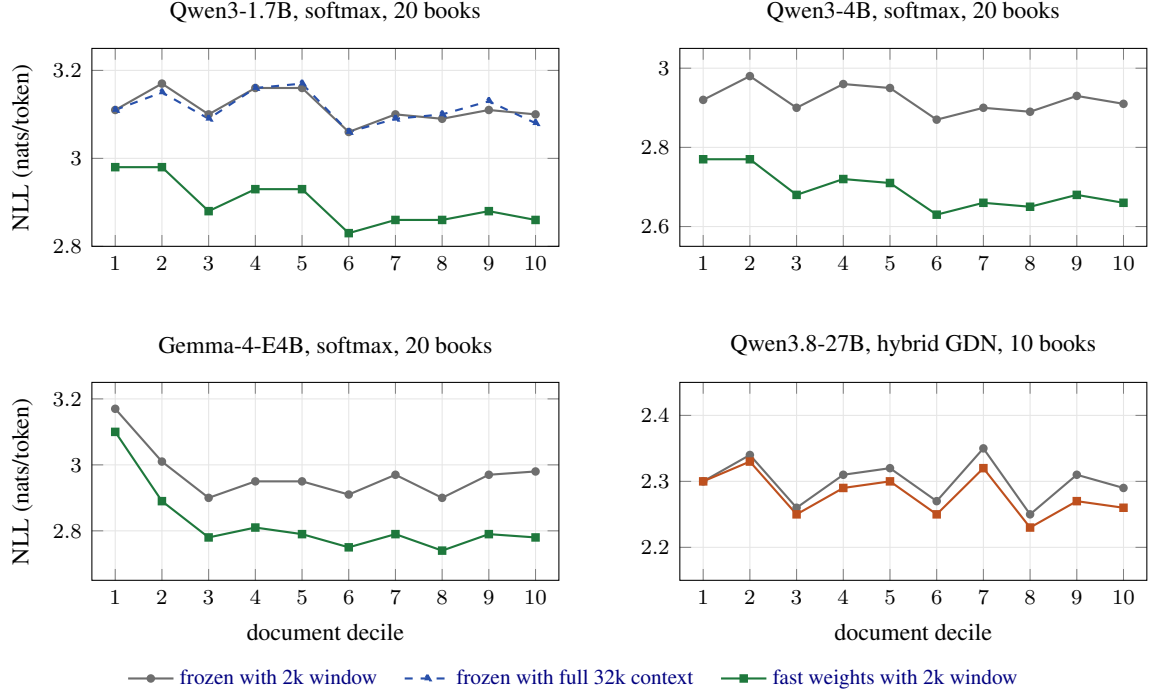
\begin{figure}[t]
\centering
\begin{tikzpicture}
\begin{groupplot}[group style={group size=2 by 2, horizontal sep=16mm, vertical sep=18mm, xlabels at=edge bottom, ylabels at=edge left}, width=0.47\linewidth, height=4.2cm, xlabel={document decile}, ylabel={NLL (nats/token)}, xtick={1,...,10}, xmin=0.5, xmax=10.5, grid=major, grid style={gray!20}, legend cell align=left]
\nextgroupplot[title={Qwen3-1.7B, softmax, 20 books}, ymin=2.8, ymax=3.25, legend to name=declegend, legend columns=3, legend style={/tikz/every even column/.append style={column sep=8pt}}]
\addplot[cfrozen, thick, mark=*, mark size=1.2] coordinates {(1,3.11)(2,3.17)(3,3.10)(4,3.16)(5,3.16)(6,3.06)(7,3.10)(8,3.09)(9,3.11)(10,3.10)};
\addplot[cfull, thick, dashed, mark=triangle*, mark size=1.4] coordinates {(1,3.11)(2,3.15)(3,3.09)(4,3.16)(5,3.17)(6,3.06)(7,3.09)(8,3.10)(9,3.13)(10,3.08)};
\addplot[cttt, thick, mark=square*, mark size=1.2] coordinates {(1,2.98)(2,2.98)(3,2.88)(4,2.93)(5,2.93)(6,2.83)(7,2.86)(8,2.86)(9,2.88)(10,2.86)};
\legend{frozen with 2k window, frozen with full 32k context, fast weights with 2k window}
\nextgroupplot[title={Qwen3-4B, softmax, 20 books}, ymin=2.55, ymax=3.05]
\addplot[cfrozen, thick, mark=*, mark size=1.2] coordinates {(1,2.92)(2,2.98)(3,2.90)(4,2.96)(5,2.95)(6,2.87)(7,2.90)(8,2.89)(9,2.93)(10,2.91)};
\addplot[cttt, thick, mark=square*, mark size=1.2] coordinates {(1,2.77)(2,2.77)(3,2.68)(4,2.72)(5,2.71)(6,2.63)(7,2.66)(8,2.65)(9,2.68)(10,2.66)};
\nextgroupplot[title={Gemma-4-E4B, softmax, 20 books}, ymin=2.65, ymax=3.25]
\addplot[cfrozen, thick, mark=*, mark size=1.2] coordinates {(1,3.17)(2,3.01)(3,2.90)(4,2.95)(5,2.95)(6,2.91)(7,2.97)(8,2.90)(9,2.97)(10,2.98)};
\addplot[cttt, thick, mark=square*, mark size=1.2] coordinates {(1,3.10)(2,2.89)(3,2.78)(4,2.81)(5,2.79)(6,2.75)(7,2.79)(8,2.74)(9,2.79)(10,2.78)};
\nextgroupplot[title={Qwen3.8-27B, hybrid GDN, 10 books}, ymin=2.15, ymax=2.45]
\addplot[cfrozen, thick, mark=*, mark size=1.2] coordinates {(1,2.30)(2,2.34)(3,2.26)(4,2.31)(5,2.32)(6,2.27)(7,2.35)(8,2.25)(9,2.31)(10,2.29)};
\addplot[chyb, thick, mark=square*, mark size=1.2] coordinates {(1,2.30)(2,2.33)(3,2.25)(4,2.29)(5,2.30)(6,2.25)(7,2.32)(8,2.23)(9,2.27)(10,2.26)};
\end{groupplot}
\end{tikzpicture}\\[2pt]
\ref{declegend}
\caption{Loss by document decile, frozen versus fast weights, at each model's best rate. On the softmax models the adapted curve sits below the frozen curve from the first decile and the gap widens through the book; the full-context control on Qwen3-1.7B lies on top of the 2k-window frozen curve. On the hybrid 27B the two curves nearly coincide.}
\label{fig:deciles}
\end{figure}

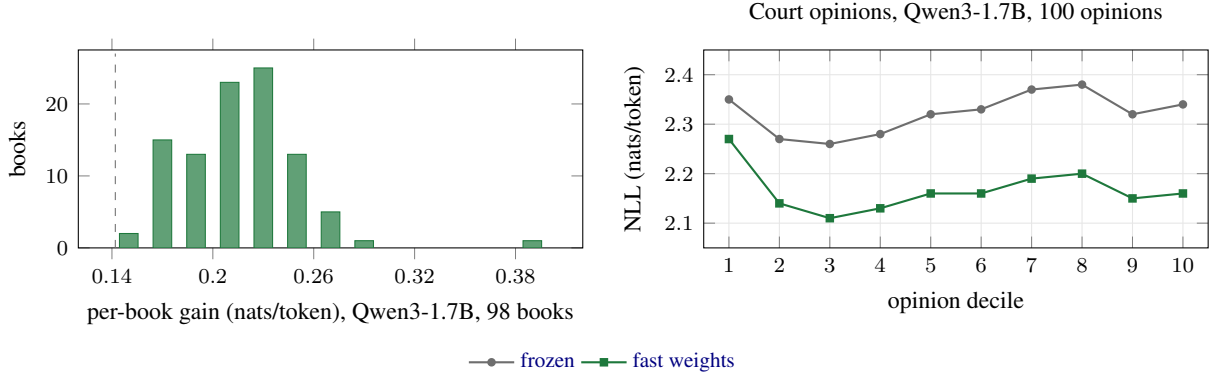
\begin{figure}[t]
\centering
\begin{tikzpicture}
\begin{groupplot}[group style={group size=2 by 1, horizontal sep=16mm}, height=4.2cm]
\nextgroupplot[width=0.5\linewidth, ybar, /pgf/bar width=7pt, xlabel={per-book gain (nats/token), Qwen3-1.7B, 98 books}, ylabel={books}, xmin=0.12, xmax=0.42, ymin=0, xtick={0.14,0.20,0.26,0.32,0.38}, x tick label style={/pgf/number format/fixed, /pgf/number format/precision=2}]
\addplot[fill=cttt!70, draw=cttt] coordinates {(0.15,2)(0.17,15)(0.19,13)(0.21,23)(0.23,25)(0.25,13)(0.27,5)(0.29,1)(0.39,1)};
\draw[dashed, cfrozen] (axis cs:0.142,0) -- (axis cs:0.142,27) node[above, font=\scriptsize] {min 0.142};
\nextgroupplot[width=0.5\linewidth, xlabel={opinion decile}, ylabel={NLL (nats/token)}, xtick={1,...,10}, xmin=0.5, xmax=10.5, ymin=2.05, ymax=2.45, grid=major, grid style={gray!20}, legend to name=legallegend, legend columns=2, legend cell align=left, title={Court opinions, Qwen3-1.7B, 100 opinions}]
\addplot[cfrozen, thick, mark=*, mark size=1.2] coordinates {(1,2.35)(2,2.27)(3,2.26)(4,2.28)(5,2.32)(6,2.33)(7,2.37)(8,2.38)(9,2.32)(10,2.34)};
\addplot[cttt, thick, mark=square*, mark size=1.2] coordinates {(1,2.27)(2,2.14)(3,2.11)(4,2.13)(5,2.16)(6,2.16)(7,2.19)(8,2.20)(9,2.15)(10,2.16)};
\legend{frozen, fast weights}
\end{groupplot}
\end{tikzpicture}\\[2pt]
\ref{legallegend}
\caption{Left: distribution of per-book gains over the 98 PG-19 books (bins of 0.02 nats); every book improves. Right: loss by decile over 100 court opinions; the gain is present from the first tenth and stable thereafter.}
\label{fig:dist}
\end{figure}

\textbf{Softmax-attention models gain 5--8\% of their loss on every document.} Table~\ref{tab:adapt} gives the main result. Figure~\ref{fig:deciles} shows the shape of the effect. We ran Qwen3-1.7B at $\eta=10^{-2}$ on the PG-19 test split. It improves on all 98 long books by $0.216$ nats per token. The minimum gain is $0.142$ and the maximum is $0.396$ (Figure~\ref{fig:dist}, left). Qwen3-4B, the same architecture at a larger size, gains $0.227$ on 20 books. Gemma-4-E4B, a different family, gains $0.145$. On court opinions the base loss is lower (2.3 vs 3.1 nats). The two Qwen3 models gain $0.15$ nats on all 100 of 100 opinions. Per-opinion gains stay between $0.11$ and $0.20$ (Figure~\ref{fig:dist}, right). We checked the decile curves. The frozen loss runs flat across the document, and the adapted loss is already lower in the first tenth, 3.11 vs 2.98 at $\eta=10^{-2}$. It keeps falling to the last tenth, 3.10 vs 2.86, a 16\% drop in perplexity by the end of a book.

\textbf{The gain exceeds what the full 32k context gives.} We ran Qwen3-1.7B on the same 20 books. The frozen model kept its native 32k window and scored 3.091 nats. We clipped the frozen model to 2k and scored 3.093. With fast weights at the 2k window, we scored 2.922. In Figure~\ref{fig:deciles} the full-context curve lies on top of the 2k-window curve at every decile. Attention over the whole book added nothing to these books; the gradient steps removed 0.17 nats. The two mechanisms extract different things. Attention copies; the gradient reweights the output distribution toward the document's vocabulary and register.

\begin{figure}[t]
\centering
\begin{tikzpicture}
\begin{axis}[width=0.72\linewidth, height=5.6cm, xmode=log, xlabel={learning rate $\eta$}, ylabel={gain (nats/token)}, xmin=2e-4, xmax=5e-2, ymin=-0.08, ymax=0.26, grid=major, grid style={gray!20}, legend pos=outer north east, legend cell align=left, log ticks with fixed point]
\addplot[cttt, thick, mark=*, mark size=1.5] coordinates {(3e-4,0.121)(1e-3,0.171)(3e-3,0.203)(1e-2,0.219)(3e-2,0.209)};
\addplot[cttt!60!black, thick, mark=square*, mark size=1.5] coordinates {(1e-3,0.156)(1e-2,0.227)};
\addplot[cttt!50!cfull, thick, mark=triangle*, mark size=1.8] coordinates {(3e-4,0.143)(1e-3,0.145)};
\addplot[chyb, thick, mark=diamond*, mark size=1.8] coordinates {(3e-4,0.006)(1e-3,0.012)(3e-3,0.012)(1e-2,-0.05)};
\addplot[chyb!60!black, thick, mark=pentagon*, mark size=1.8] coordinates {(1e-3,0.008)(1e-2,0.019)};
\addplot[cfrozen, dashed] coordinates {(2e-4,0)(5e-2,0)};
\legend{Qwen3-1.7B (softmax), Qwen3-4B (softmax), Gemma-4-E4B (softmax), Qwen3.5-4B (hybrid), Qwen3.8-27B (hybrid), no change}
\end{axis}
\end{tikzpicture}
\caption{Gain versus learning rate on PG-19 for each model. The three softmax models rise to $0.15$--$0.23$ nats; the two hybrid models stay within $0.02$ of zero at every rate and Qwen3.5-4B turns negative at $10^{-2}$ (plotted at its mean over 12 books).}
\label{fig:lr}
\end{figure}
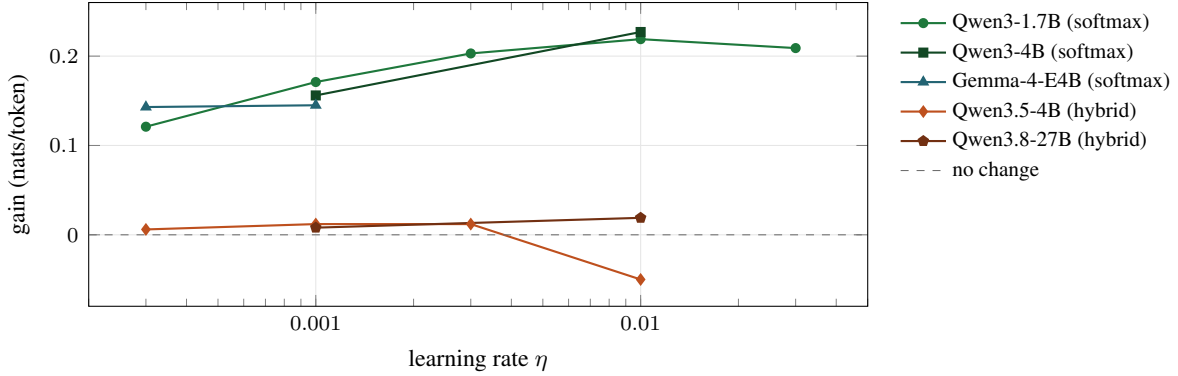

\textbf{Hybrid linear-attention models gain under 1\%.} Figure~\ref{fig:lr} shows the per-model rate curves. We ran Qwen3.5-4B at the softmax models' rate of $10^{-2}$. It got worse on 12 of 12 books, $-0.02$ to $-0.08$ nats, drift 1.0\%. At $3\times10^{-4}$, $10^{-3}$ and $3\times10^{-3}$ it gained $0.006$, $0.012$ and $0.012$ nats. The gains were statistically positive on 20 of 20 books and practically absent. Qwen3.8-27B, the same hybrid design at 27B, gained $0.008$ at $10^{-3}$ and $0.019$ at $10^{-2}$. The contrast with Qwen3-4B isolates the attention architecture as the cause. Qwen3-4B shares the training generation, tokenizer and size. Gated DeltaNet blocks already carry a delta-rule fast-weight state that adapts to the document in context. A late-layer gradient step has little left to add.

\subsection{Masked-diffusion language models}

\begin{table}[t]
\caption{Adaptation on masked-diffusion language models. Loss is the model's own denoising objective ($1/t$-weighted masked cross-entropy; ``plain CE'' unweighted), comparable within a model. 16k tokens per document, chunk 256, window 2k, update clipping 0.5.}
\label{tab:diffusion}
\centering\small
\begin{tabular}{llrrrrrc}
\toprule
Model & Stream & $n$ & $\eta$ & Frozen loss & Gain $\uparrow$ & Rel. & Improved \\
\midrule
Nemotron-Diffusion-8B & PG-19 & 10 & $3\times10^{-4}$ & 10.6 & \best{$+0.82$} & 8\% & 10/10 \\
Nemotron-Diffusion-8B & PG-19 & 10 & $10^{-3}$ & 10.6 & $+0.67$ & 6\% & 8/10 \\
Nemotron-Diffusion-8B, plain CE & PG-19 & 10 & $10^{-3}$ & 4.7 & $+0.31$ & 7\% & 10/10 \\
Nemotron-Diffusion-8B & opinions & 20 & $10^{-3}$ & 7.5 & $+0.50$ & 7\% & 18/20 \\
DiffusionGemma-26B-A4B & PG-19 & 10 & $10^{-3}$ & 11.9 & \best{$+2.42$} & 20\% & 10/10 \\
DiffusionGemma-26B-A4B & opinions & 20 & $10^{-3}$ & 8.5 & \best{$+1.47$} & 17\% & 20/20 \\
\bottomrule
\end{tabular}
\end{table}

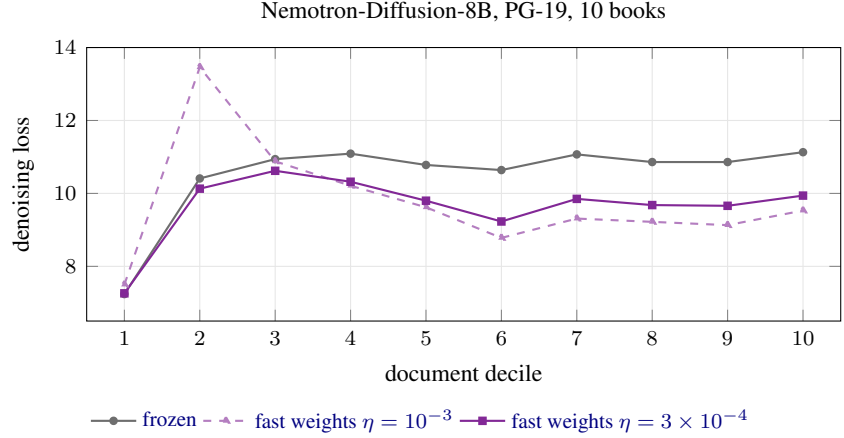
\begin{figure}[t]
\centering
\begin{tikzpicture}
\begin{axis}[width=0.7\linewidth, height=5.2cm, xlabel={document decile}, ylabel={denoising loss}, xtick={1,...,10}, xmin=0.5, xmax=10.5, ymin=6.5, ymax=14, grid=major, grid style={gray!20}, legend to name=difflegend, legend columns=3, legend cell align=left, title={Nemotron-Diffusion-8B, PG-19, 10 books}]
\addplot[cfrozen, thick, mark=*, mark size=1.2] coordinates {(1,7.24)(2,10.41)(3,10.94)(4,11.09)(5,10.78)(6,10.64)(7,11.07)(8,10.86)(9,10.86)(10,11.13)};
\addplot[cdiff!60, thick, dashed, mark=triangle*, mark size=1.4] coordinates {(1,7.50)(2,13.46)(3,10.88)(4,10.21)(5,9.62)(6,8.78)(7,9.31)(8,9.22)(9,9.13)(10,9.53)};
\addplot[cdiff, thick, mark=square*, mark size=1.2] coordinates {(1,7.26)(2,10.13)(3,10.62)(4,10.32)(5,9.80)(6,9.23)(7,9.85)(8,9.68)(9,9.66)(10,9.94)};
\legend{frozen, fast weights $\eta=10^{-3}$, fast weights $\eta=3\times10^{-4}$}
\end{axis}
\end{tikzpicture}\\[2pt]
\ref{difflegend}
\caption{Denoising loss by decile on a diffusion LM. At $10^{-3}$ the second decile overshoots before the gain opens; at $3\times10^{-4}$ the adapted curve never rises above the frozen curve and ends 11\% lower. The first decile is low for both because the first chunks of a book are short front matter.}
\label{fig:diffusion}
\end{figure}

\textbf{The same gradient step lowers the denoising loss of two diffusion families, at a lower rate.} We ran the first measurement of gradient adaptation on masked-diffusion LMs. Table~\ref{tab:diffusion} reports it and Figure~\ref{fig:diffusion} shows the decile shape. At the autoregressive models' rate of $10^{-2}$ without clipping, Nemotron-Diffusion-8B diverged on the first book. Loss went 9.3 to 41.6, drift 11\%. The $1/t$-weighted objective has much larger gradients. At $3\times10^{-4}$ with clipping it gained 8\% on every book, no early overshoot, and a monotone decile curve; at $10^{-3}$ the first two deciles overshot before a 14\% gain by the last. The unweighted objective gave the same 7\% relative gain, so the rate and not the weighting is what matters. DiffusionGemma-26B, through its context/canvas interface, gained 20\% on books and 17\% on opinions, on every document. Its decile curve fell from 12.1 to 10.7 in the first tenth, from 11.9 to 9.1 in the last. Both diffusion models are softmax-attention denoisers. The direction and the decile shape match the autoregressive softmax models.

\subsection{What controls the gain}
\label{sec:controls}

\begin{table}[t]
\caption{Ablations on Qwen3-1.7B, 20 PG-19 books, 32k tokens, window 2k. Gain in nats per token relative to the frozen model; the reference configuration is one SGD step at $\eta=10^{-2}$ on the last quarter of blocks with a constant rate.}
\label{tab:ablate}
\centering\small
\begin{tabular}{llrl}
\toprule
Factor & Setting & Gain $\uparrow$ & Note \\
\midrule
\multirow{5}{*}{learning rate} & $3\times10^{-4}$ & $+0.121$ & \\
 & $10^{-3}$ & $+0.171$ & \\
 & $3\times10^{-3}$ & $+0.203$ & \\
 & $10^{-2}$ & \best{$+0.219$} & \\
 & $3\times10^{-2}$ & $+0.209$ & first decile worse than frozen \\
\midrule
\multirow{3}{*}{writable blocks ($\eta=10^{-2}$)} & first quarter (0--6) & $-0.032$ & $\pm 0.044$, 9/20 improve \\
 & middle quarter (10--16) & $+0.182$ & $\pm 0.014$, 20/20 \\
 & last quarter (21--27) & \best{$+0.219$} & 20/20 \\
\midrule
\multirow{3}{*}{rate gate ($\eta=10^{-3}$)} & constant & \best{$+0.171$} & \\
 & sigmoid of surprise $z$-score & $+0.144$ & \\
 & clipped surprise & $+0.144$ & \\
\midrule
\multirow{2}{*}{inner steps ($\eta=10^{-2}$)} & 1 & \best{$+0.219$} & \\
 & 4 with update clipping 0.5 & $+0.183$ & \\
\bottomrule
\end{tabular}
\end{table}

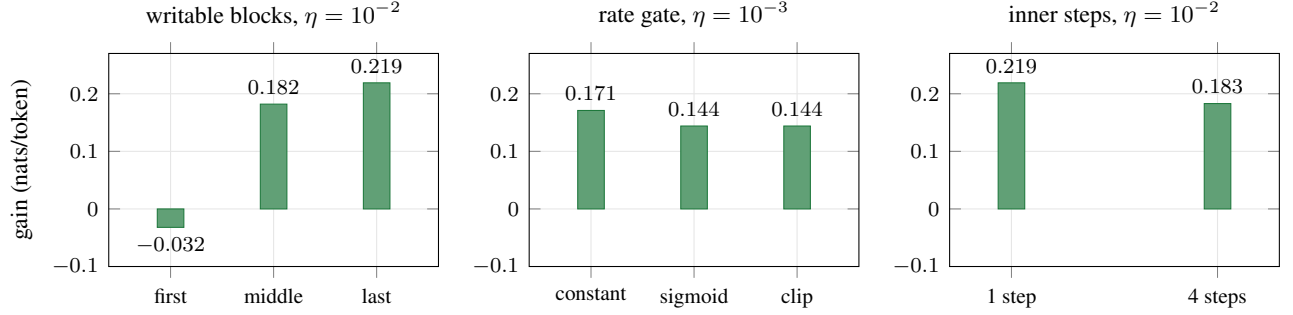
\begin{figure}[t]
\centering
\begin{tikzpicture}
\begin{groupplot}[group style={group size=3 by 1, horizontal sep=12mm}, height=4.4cm, width=0.36\linewidth, ybar, /pgf/bar width=10pt, ymin=-0.1, ymax=0.27, ylabel={gain (nats/token)}, grid=major, grid style={gray!20}, nodes near coords, every node near coord/.append style={font=\scriptsize, /pgf/number format/fixed, /pgf/number format/precision=3}, xtick=data, x tick label style={font=\scriptsize, align=center}, enlarge x limits=0.3]
\nextgroupplot[title={writable blocks, $\eta=10^{-2}$}, symbolic x coords={first, middle, last}]
\addplot[fill=cttt!70, draw=cttt] coordinates {(first,-0.032)(middle,0.182)(last,0.219)};
\nextgroupplot[title={rate gate, $\eta=10^{-3}$}, symbolic x coords={constant, sigmoid, clip}, ylabel={}]
\addplot[fill=cttt!70, draw=cttt] coordinates {(constant,0.171)(sigmoid,0.144)(clip,0.144)};
\nextgroupplot[title={inner steps, $\eta=10^{-2}$}, symbolic x coords={1 step, 4 steps}, ylabel={}]
\addplot[fill=cttt!70, draw=cttt] coordinates {(1 step,0.219)(4 steps,0.183)};
\end{groupplot}
\end{tikzpicture}
\caption{Ablations on Qwen3-1.7B, 20 books. Late blocks, a constant rate and a single step are each the best setting of their factor.}
\label{fig:ablate}
\end{figure}

\textbf{One large step on late blocks with a constant rate is the best configuration, and each alternative loses.} Table~\ref{tab:ablate} and Figure~\ref{fig:ablate} collect the ablations. The gain rises with the rate up to $10^{-2}$ and at $3\times10^{-2}$ the first decile overshoots (3.14 against 3.11 frozen). Early blocks are useless and slightly harmful; middle blocks capture 83\% of the late-block gain, placing the adaptation on the output side of the network. A surprise-dependent gate, whether a sigmoid of the loss $z$-score or a clipped version, gives the gain of a constant rate at about half the value, because its mean is about 0.5 and the shape of token surprise carries nothing: the most surprising tokens in a court opinion are docket numbers. Four inner steps with clipping lose to one step. Each of these results repeats on the court-opinion stream at 40 opinions: constant $0.099$ versus gated $0.076$ at $10^{-3}$, and one step $0.155$ versus four steps $0.119$ at $10^{-2}$.

\subsection{Retrieval from the updated weights}
\label{sec:retrieval}

\begin{table}[t]
\caption{Retrieval benchmarks under the three-condition protocol, $\eta=10^{-2}$, chunk 512, window 2k. Full = frozen model with the whole document; window = frozen model with the last 2k tokens; FW = fast weights updated on the whole document, then answering from the window. Accuracy unless marked; RULER uses the strict all-answers criterion; $n$ items per cell. Context-truncation loss is Full $-$ Window; recovery is FW $-$ Window.}
\label{tab:retrieval}
\centering\footnotesize\setlength{\tabcolsep}{3.5pt}
\begin{tabular}{lrrrrrrrr}
\toprule
& \multicolumn{4}{c}{Qwen3.8-27B} & \multicolumn{4}{c}{Qwen3-4B}\\
\cmidrule(lr){2-5}\cmidrule(lr){6-9}
Task & $n$ & Full & Window & FW & $n$ & Full & Window & FW \\
\midrule
RULER 4k & 100 & 0.87 & 0.40 & 0.38 & 100 & 0.84 & 0.32 & 0.31 \\
RULER 16k & & & & & 100 & 0.81 & 0.18 & 0.18 \\
BABILong 16k qa1 & 100 & 0.67 & 0.23 & 0.23 & 100 & 0.39 & 0.11 & 0.12 \\
LongBench-v2 & 149 & 0.64$^\dagger$ & 0.42 & 0.44 & & & & \\
SCOTUS issue area (14-way) & 100 & 0.65 & 0.40 & 0.40 & 100 & 0.57 & 0.53 & 0.53 \\
ContractNLI (3-way) & 100 & 0.69 & 0.61 & 0.60 & 100 & 0.54 & 0.54 & 0.56 \\
Citation completion, exact & 177 & 0.25 & 0.21 & 0.20 & 158 & 0.15 & 0.11 & 0.11 \\
HotpotQA F1 (FW: passages read first) & 100 & 0.81 & & 0.80 & 100 & 0.31 & & 0.31 \\
MuSiQue F1 (FW: passages read first) & 100 & 0.67 & & 0.67 & 100 & 0.22 & & 0.22 \\
CounterFact (Full = fact in prompt) & 100 & 0.76 & 0.01 & 0.04 & 100 & 0.76 & 0.02 & 0.02 \\
\bottomrule
\multicolumn{9}{l}{$^\dagger$Full context on the 56 of 60 items of the first LongBench-v2 run that fit in memory.}
\end{tabular}
\end{table}

\begin{figure}[t]
\centering
\begin{tikzpicture}
\begin{axis}[width=0.9\linewidth, height=5.6cm, ybar, /pgf/bar width=7pt, ymin=0, ymax=1.0, ylabel={accuracy}, symbolic x coords={RULER 4k, BABILong 16k, LongBench-v2, SCOTUS, ContractNLI, Citation exact}, xtick=data, x tick label style={font=\scriptsize}, grid=major, grid style={gray!20}, legend to name=retrlegend, legend columns=3, legend cell align=left, enlarge x limits=0.12, title={Qwen3.8-27B, $n \ge 100$ per task}]
\addplot[fill=cfull!60, draw=cfull] coordinates {(RULER 4k,0.87)(BABILong 16k,0.67)(LongBench-v2,0.64)(SCOTUS,0.65)(ContractNLI,0.69)(Citation exact,0.25)};
\addplot[fill=cfrozen!40, draw=cfrozen] coordinates {(RULER 4k,0.40)(BABILong 16k,0.23)(LongBench-v2,0.42)(SCOTUS,0.40)(ContractNLI,0.61)(Citation exact,0.21)};
\addplot[fill=cttt!70, draw=cttt] coordinates {(RULER 4k,0.38)(BABILong 16k,0.23)(LongBench-v2,0.44)(SCOTUS,0.40)(ContractNLI,0.60)(Citation exact,0.20)};
\legend{frozen with full context, frozen with 2k window, fast weights + 2k window}
\end{axis}
\end{tikzpicture}\\[2pt]
\ref{retrlegend}
\caption{The retrieval result on the 27B in one picture. The gap between the first and second bar of each task is what truncation to the window costs; the third bar is what the updated weights give back, which is nothing on every task.}
\label{fig:retrieval}
\end{figure}
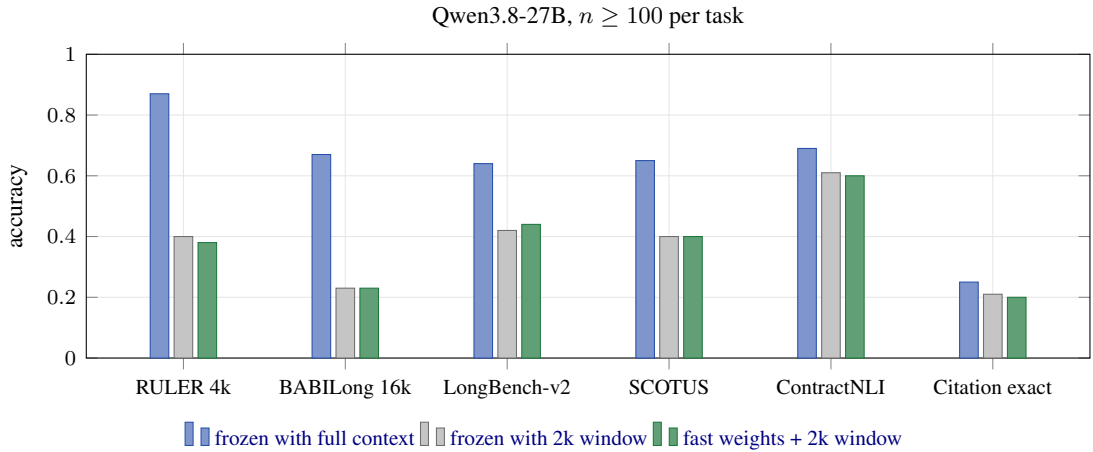

\textbf{Weights updated on the full document recover none of the accuracy lost to the window.} Table~\ref{tab:retrieval} and Figure~\ref{fig:retrieval} cover needle retrieval, multi-hop reasoning, long-document classification, long-contract entailment, in-document citation and single-fact recall, on a 4B softmax model and a 27B hybrid model. Truncating the context costs the 27B 47 points on RULER 4k, 44 on BABILong, 22 on LongBench-v2, 25 on SCOTUS and 8 on ContractNLI. In every case the fast-weight condition is within two points of the window condition. The one apparent positive, LongBench-v2 at $n=60$ where fast weights recovered 5 points, shrank to 1.3 points at $n=149$ with a different seed. On multi-hop QA, reading the passages into the weights before answering with them in context changes F1 by under half a point. A single fast-weight step on a new fact leaves the model at its no-fact baseline. Demonstrations fare the same: learning four CaseHOLD demonstrations into the weights, for one or ten passes, leaves zero-shot accuracy at the zero-shot value on both models (27B: 0.73 zero-shot, 0.72 after the write; 4B: 0.58 and 0.58).

\textbf{The negative is systematic, not a scorer artifact.} We ran an earlier version of the choice-task scorer first, one that compared option-letter probabilities after a chat template's assistant turn. The letters sit at $\log p \approx -40$ there. That scorer produced a 3-of-7 recovery on SCOTUS. With raw completion scoring the SCOTUS truncation gap is 25 points and the recovery is zero. We kept the raw scorer for every choice task in Table~\ref{tab:retrieval}.

\subsection{Writing facts: format, not knowledge}
\label{sec:facts}

\begin{table}[t]
\caption{Fact writes by the next-token objective and their cost. Top: recall of a written fact. Bottom: Qwen3-4B before and after a session of 30 sequential fact writes (20 passes each), and Qwen3-8B before and after one behaviour write, on 200-item retention probes. Drift is the relative change of $\fw$.}
\label{tab:facts}
\centering\footnotesize\setlength{\tabcolsep}{4pt}
\begin{tabular}{llrrr}
\toprule
\multicolumn{5}{l}{\textit{Recall of written facts}} \\
Setting & Prompt form & In prompt & In weights, verbatim & In weights, rephrased \\
\midrule
CounterFact, Qwen3-4B, 30 facts, 20 passes at $10^{-2}$ & raw & 0.83 & 1.00 & 0.50 \\
zsRE, Qwen3.8-27B, 100 edits, 20 passes at $10^{-2}$ & raw Q/A & 0.53 & 1.00 & 0.78 \\
zsRE, same 100 edits and weights & chat template & 0.80 & 0.30 & 0.30 \\
\midrule
\multicolumn{5}{l}{\textit{Retention after a session of writes}} \\
Write & Drift & GSM8K strict & GSM8K flexible & MMLU / IFEval \\
\midrule
none (Qwen3-4B base) & & 0.865 & 0.86 & 0.715 / 0.185 \\
30 facts, next-token, $\eta=10^{-2}$ & 0.48\% & 0.485 & 0.72 & 0.722 / 0.150 \\
30 facts, next-token, $\eta=2\times10^{-3}$ & 0.32\% & 0.50 & 0.755 & \\
\midrule
none (Qwen3-8B base) & & 0.915 & 0.92 & \\
one rule, behaviour write & 0.58\% & 0.90 & 0.91 & \\
\bottomrule
\end{tabular}
\end{table}

\begin{figure}[t]
\centering
\begin{tikzpicture}
\begin{groupplot}[group style={group size=2 by 1, horizontal sep=16mm}, height=4.6cm, width=0.5\linewidth, ybar, /pgf/bar width=9pt, ymin=0, ymax=1.05, grid=major, grid style={gray!20}, xtick=data, x tick label style={font=\scriptsize, align=center}, enlarge x limits=0.25, nodes near coords, every node near coord/.append style={font=\scriptsize, /pgf/number format/fixed, /pgf/number format/precision=2}]
\nextgroupplot[title={CounterFact recall, Qwen3-4B}, ylabel={recall}, symbolic x coords={1 step, {20 passes\\ verbatim}, {20 passes\\ rephrased}, in prompt}]
\addplot[fill=cfrozen!40, draw=cfrozen] coordinates {(1 step,0.03)({20 passes\\ verbatim},1.00)({20 passes\\ rephrased},0.50)(in prompt,0.83)};
\nextgroupplot[title={GSM8K strict after 30 writes}, ylabel={accuracy}, symbolic x coords={base, {$\eta=2\!\times\!10^{-3}$}, {$\eta=10^{-2}$}, {behaviour\\ write (8B)}}]
\addplot[fill=chyb!50, draw=chyb] coordinates {(base,0.865)({$\eta=2\!\times\!10^{-3}$},0.50)({$\eta=10^{-2}$},0.485)({behaviour\\ write (8B)},0.90)};
\end{groupplot}
\end{tikzpicture}
\caption{Left: a fact is stored only at a high dose and mostly in its training form. Right: both storing doses of the next-token write halve GSM8K; the behaviour write, with larger drift, leaves it unchanged (8B base 0.915).}
\label{fig:facts}
\end{figure}
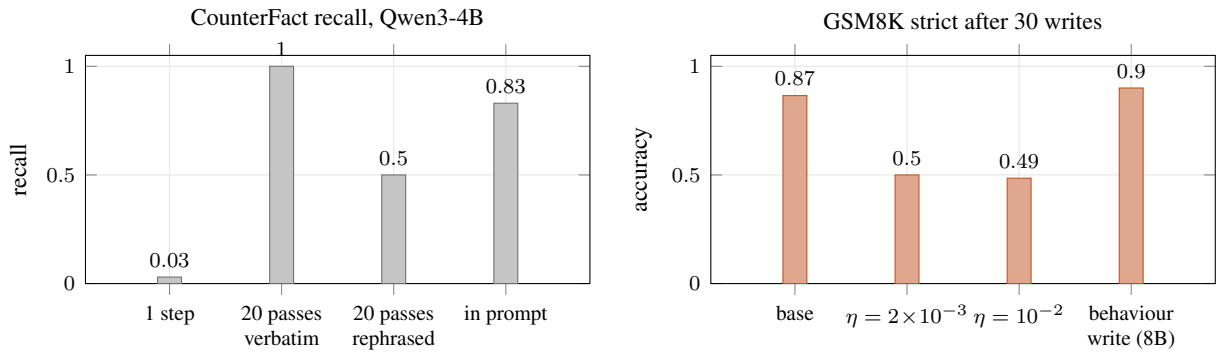

\textbf{Next-token writes store the training format.} Table~\ref{tab:facts} (top) and Figure~\ref{fig:facts} (left) show the dose at which a fact can be stored at all: one step changes nothing (recall 0.03), and 20 passes at $\eta=10^{-2}$ give verbatim recall of 1.00 and 0.50 on a rephrased question on CounterFact. On zsRE with the 27B, the same dose gives 0.78 on rephrased questions when both the write and the test use a raw \texttt{Question/Answer} layout, and 0.30 when the test question goes through the chat template, against 0.80 with the fact in the prompt. The write taught a completion pattern; the fact is available only in that pattern.

\textbf{The storing dose breaks generated reasoning at every rate tried.} Table~\ref{tab:facts} (bottom) and Figure~\ref{fig:facts} (right): after 30 sequential writes at $\eta=10^{-2}$, which yield 0.23 rephrased recall across the session, MMLU is unchanged while GSM8K strict falls from 0.865 to 0.485 and IFEval loses 4 points. Of the 38-point strict loss, 24 points are format collapse (94 of 200 generations omit the answer marker) and 14 points are reasoning loss on the lenient metric (0.86 to 0.72). A fifth of the rate stores less than half as much (0.10) and still costs 36 strict and 11 lenient points. Damage is not proportional to dose. The behaviour write in the last row moves the same parameter set by a larger relative amount, learns its rule to AUROC 1.00, and changes GSM8K by 1--1.5 points inside the noise of 200 items. The objective, not the amount of weight change, decides whether general ability survives.

\subsection{The inquiry loop}
\label{sec:inquiryresults}

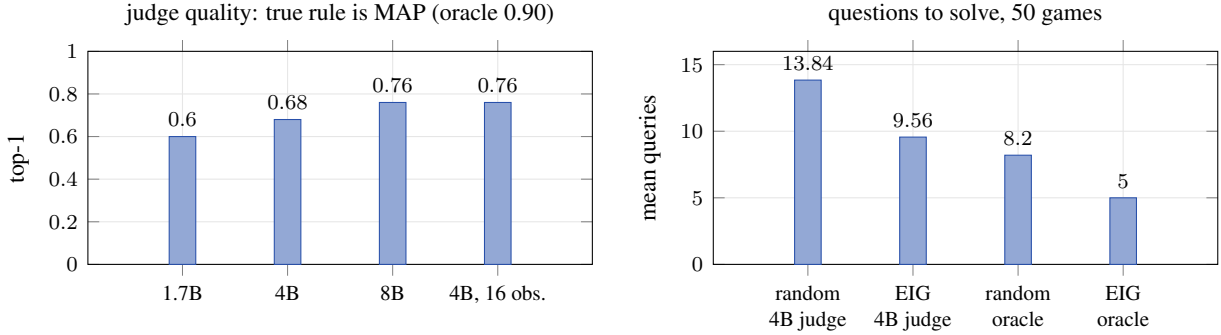
\begin{figure}[t]
\centering
\begin{tikzpicture}
\begin{groupplot}[group style={group size=2 by 1, horizontal sep=16mm}, height=4.4cm, width=0.5\linewidth, ybar, /pgf/bar width=10pt, grid=major, grid style={gray!20}, xtick=data, x tick label style={font=\scriptsize, align=center}, enlarge x limits=0.3, nodes near coords, every node near coord/.append style={font=\scriptsize, /pgf/number format/fixed, /pgf/number format/precision=2}]
\nextgroupplot[title={judge quality: true rule is MAP (oracle 0.90)}, ylabel={top-1}, ymin=0, ymax=1.0, symbolic x coords={1.7B, 4B, 8B, {4B, 16 obs.}}]
\addplot[fill=cfull!50, draw=cfull] coordinates {(1.7B,0.60)(4B,0.68)(8B,0.76)({4B, 16 obs.},0.76)};
\nextgroupplot[title={questions to solve, 50 games}, ylabel={mean queries}, ymin=0, ymax=16, symbolic x coords={{random\\ 4B judge}, {EIG\\ 4B judge}, {random\\ oracle}, {EIG\\ oracle}}]
\addplot[fill=cfull!50, draw=cfull] coordinates {({random\\ 4B judge},13.84)({EIG\\ 4B judge},9.56)({random\\ oracle},8.2)({EIG\\ oracle},5.0)};
\end{groupplot}
\end{tikzpicture}
\caption{The inquiry loop. Left: how often the posterior's mode is the true rule after 8 observations, by judge size; 16 observations lift the 4B to the 8B's level. Right: information-gain selection removes 31\% of the questions with the 4B judge and 39\% with exact likelihoods.}
\label{fig:inquiry}
\end{figure}

\textbf{The loop's quality is the judge's per-observation accuracy, and information-gain selection pays.} Figure~\ref{fig:inquiry} summarises the loop we ran on 50 games with 8 observations each. With the few-shot likelihood prompt, the true rule was the posterior mode in 60\% of games for the 1.7B judge, 68\% for the 4B and 76\% for the 8B, against an oracle of 90\%. Calibration error for the 8B was 0.098. The judges' per-observation accuracies came in at 0.77, 0.86 and about 0.90, and the posterior tracked them. We gave the 4B 16 observations instead of 8 and it lifted to 0.76 against a 0.92 oracle. Temperature and per-hypothesis bias correction did not move top-1, because the error was noise rather than bias. Expected-information-gain question selection solved the games in 9.56 questions against 13.84 for random selection with the 4B judge. With the oracle likelihood, it solved the games in 5.0 against 8.2. Disagreement-based selection picked the same questions as EIG on binary answers.

\subsection{Writing conclusions as behaviour}
\label{sec:behaviour}

\begin{table}[t]
\caption{Writing a conclusion from the inquiry loop into the weights. Hidden-rule environment, balanced held-out cases scored against the true rule, no rule in the prompt at test time. All entries are AUROC against the true rule on held-out cases. ``Ceiling'' is the same model with the rule in the prompt; ``collateral'' is AUROC at following a different rule given in the prompt, before $\rightarrow$ after the write. Teacher labels are the model's own judgments with the rule in the prompt.}
\label{tab:behaviour}
\centering\footnotesize\setlength{\tabcolsep}{4pt}
\begin{tabular}{llrrrrr}
\toprule
Model & Write & $n$ games & Chance & Ceiling & Written & Collateral \\
\midrule
Qwen3-8B & text write of the rule, 20 passes, $\eta=10^{-2}$ & 30 & 0.52 & 0.85 & 0.50 & \\
Qwen3-8B & text write of rule + gathered evidence & 30 & 0.52 & 0.85 & 0.50 & \\
\midrule
Qwen3-8B & behaviour, late MLP, Adam $2\times10^{-5}$, teacher & 20 & 0.59 & 0.88 & 0.86 & $0.77\rightarrow0.73$ \\
Qwen3-8B & behaviour, late MLP, oracle labels & 20 & 0.59 & 0.88 & 0.97 & \\
Qwen3-8B & behaviour, LoRA r16, Adam $2\times10^{-4}$, oracle & 20 & 0.59 & 0.88 & 0.995 & $0.77\rightarrow0.59$ \\
Qwen3-8B & behaviour, LoRA r16, Adam $5\times10^{-5}$, teacher & 100 & 0.54 & 0.84 & \best{0.83} & $0.82\rightarrow0.81$ \\
Qwen3.5-9B & behaviour, LoRA r16, Adam $5\times10^{-5}$, teacher & 100 & 0.53 & 0.88 & \best{0.88} & $0.85\rightarrow0.83$ \\
\bottomrule
\end{tabular}
\end{table}

\begin{figure}[t]
\centering
\begin{tikzpicture}
\begin{axis}[width=0.66\linewidth, height=5.6cm, xlabel={collateral loss (other-rule prompt-following, before $-$ after)}, ylabel={written behaviour (AUROC, no rule in prompt)}, xmin=-0.01, xmax=0.2, ymin=0.6, ymax=1.02, xtick={0,0.05,0.1,0.15,0.2}, x tick label style={/pgf/number format/fixed}, grid=major, grid style={gray!20}, legend pos=outer north east, legend cell align=left, title={Qwen3-8B, 20 games per point}]
\addplot[only marks, mark=*, mark size=2.2, cttt] coordinates {(0.01,0.68)(0.04,0.86)};
\addplot[only marks, mark=square*, mark size=2.2, cfull] coordinates {(0.01,0.87)(0.08,0.88)(0.18,0.995)};
\addplot[cfrozen, dashed] coordinates {(-0.01,0.88)(0.2,0.88)};
\addplot[cfrozen, dotted] coordinates {(-0.01,0.59)(0.2,0.59)};
\node[font=\scriptsize, anchor=west] at (axis cs:0.016,0.68) {late MLP $5\!\times\!10^{-6}$};
\node[font=\scriptsize, anchor=north west] at (axis cs:0.044,0.852) {late MLP $2\!\times\!10^{-5}$};
\node[font=\scriptsize, anchor=south west] at (axis cs:0.014,0.885) {LoRA $5\!\times\!10^{-5}$};
\node[font=\scriptsize, anchor=south west] at (axis cs:0.084,0.893) {LoRA $2\!\times\!10^{-4}$, teacher};
\node[font=\scriptsize, anchor=north east] at (axis cs:0.18,0.99) {LoRA $2\!\times\!10^{-4}$, oracle};
\legend{late-MLP write, LoRA write, in-context ceiling, no rule}
\end{axis}
\end{tikzpicture}
\caption{Dose sets the trade-off between transfer and collateral. Gentle LoRA reaches the ceiling at 0.01 collateral; four times the rate buys 0.1 more transfer at 0.18 collateral.}
\label{fig:dose}
\end{figure}
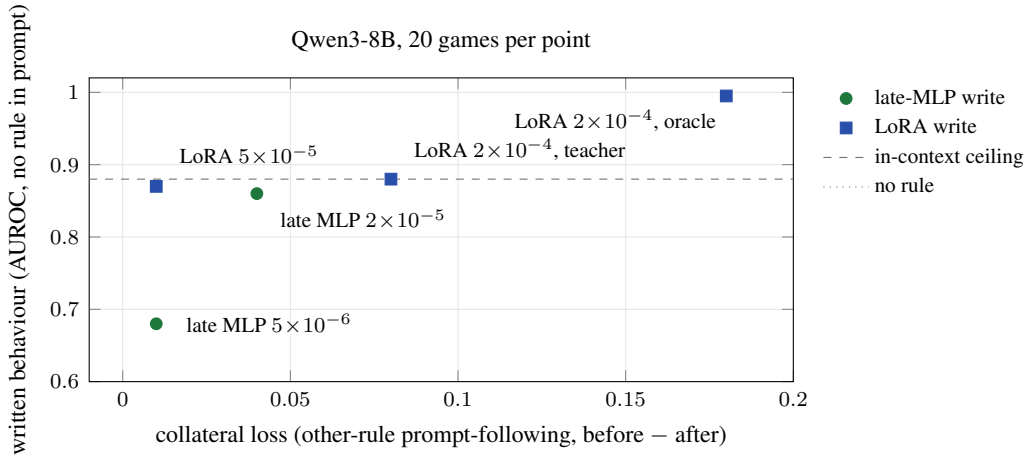

\textbf{A conclusion written as text is inert; written as behaviour it reaches the in-context ceiling.} We ran the hidden-rule environment. The inquiry loop found the true rule in every game with the oracle likelihood, at 5.1 questions. With the 8B judge it found the rule in 80\% of games, at 9.9 questions. Table~\ref{tab:behaviour} then covers writing the found rule into the weights. We ran the text write of the rule statement, with or without the evidence the loop gathered. Classification AUROC on new cases came in at 0.50, at chance. We checked that the model recited the statement perfectly afterwards. Next we ran the behaviour write of the same rule on the same parameters. It scored AUROC 0.86 against the 0.88 ceiling with teacher labels and 0.97 with oracle labels. At 100 games the gentle LoRA write landed within 0.01 of the ceiling on a softmax model (0.827 vs 0.839) and a hybrid model (0.876 vs 0.879). Unrelated prompt-following took 1--3 points of collateral. Figure~\ref{fig:dose} shows that the dose sets the collateral. LoRA at four times the rate learned the rule to 0.995. It dropped 18 points of the ability to follow a different rule from the prompt. The collateral probe is the gate that a write must pass.

\textbf{The write transfers to hybrid models at full strength, with a per-model dose.} Qwen3.5-4B reaches 0.96 with oracle labels and 0.86 with its own labels, the same as the softmax 8B; at the softmax models' rate its weights move three times more (1.8\% vs 0.6\%) and the collateral is 0.14, and at a quarter of the rate the write is 0.79 with collateral 0.01. Gentle LoRA on Qwen3.5-9B gives 0.92 against a 0.92 ceiling with collateral $0.834\rightarrow0.827$.

\textbf{On a real legal task the same write scales with labelled data.} On CaseHOLD with Qwen3-8B (200 test items), zero-shot accuracy is 0.690 and few-shot 0.665, so the model's own labels have nothing to add and the teacher write lands at 0.655. With true labels, 256 examples through the late MLP projections give 0.670 and through LoRA 0.745; 1024 examples through LoRA give 0.775, $+8.5$ points. Capacity location matters for real tasks: a rule over three integers fits in the output projections, a legal-holding judgment needs the adapter across all layers.

\section{Discussion}
\label{sec:discussion}

\textbf{Why prediction improves and retrieval does not.} A next-token gradient on a late MLP output projection moves the model's output prior toward the token statistics of the recent text: vocabulary, names as frequent tokens, formatting, register. That lowers the loss on every subsequent token of the same document, which is what the decile curves show, including the large gain in the first tenth. Attention over 32k tokens can copy but does not reweight the output distribution, which is why the full-context control gains nothing on the same books. Retrieval needs a key-to-value binding that a handful of SGD steps on a late projection does not form. TTT layers, Titans, TTT-E2E and In-Place TTT get binding because the memory or the whole model is trained to use its fast weights; the untrained setting measured here is their baseline, and its retrieval value is zero.

\begin{figure}[t]
\centering
\begin{tikzpicture}[font=\scriptsize, node distance=3mm and 6mm,
  blk/.style={draw, minimum width=16mm, minimum height=5.5mm, inner sep=1pt, align=center},
  soft/.style={blk, fill=gray!15},
  lin/.style={blk, fill=chyb!25},
  fast/.style={blk, fill=hl, thick},
  den/.style={blk, fill=cdiff!15},
  arr/.style={-{Latex[length=1.6mm]}}]
\node[font=\small\bfseries] (t1) {softmax AR};
\node[soft, below=of t1] (s1) {attention};
\node[soft, below=of s1] (s2) {MLP};
\node[soft, below=of s2] (s3) {attention};
\node[fast, below=of s3] (s4) {MLP $\fw$};
\node[below=of s4, align=center] (g1) {gain 5--8\%\\ in-context state: none};
\node[font=\small\bfseries, right=26mm of t1] (t2) {hybrid GDN};
\node[lin, below=of t2] (h1) {GDN $S_t$};
\node[lin, below=of h1] (h2) {GDN $S_t$};
\node[soft, below=of h2] (h3) {attention};
\node[fast, below=of h3] (h4) {MLP $\fw$};
\node[below=of h4, align=center] (g2) {gain 0.3--0.8\%\\ $S_t$ already adapts per token};
\node[font=\small\bfseries, right=26mm of t2] (t3) {masked diffusion};
\node[den, below=of t3] (d1) {attention (bidir.)};
\node[den, below=of d1] (d2) {MLP};
\node[den, below=of d2] (d3) {attention (bidir.)};
\node[fast, below=of d3] (d4) {MLP $\fw$};
\node[below=of d4, align=center] (g3) {gain 8--20\%\\ in-context state: none};
\foreach \a/\b in {s1/s2, s2/s3, s3/s4, h1/h2, h2/h3, h3/h4, d1/d2, d2/d3, d3/d4} \draw[arr] (\a) -- (\b);
\end{tikzpicture}
\caption{Why the architecture decides. The writable projections (green) sit at the top of each stack. Only the hybrid carries a token-by-token fast-weight state $S_t$ inside its linear-attention blocks (orange), and only the hybrid shows no gain.}
\label{fig:arch}
\end{figure}
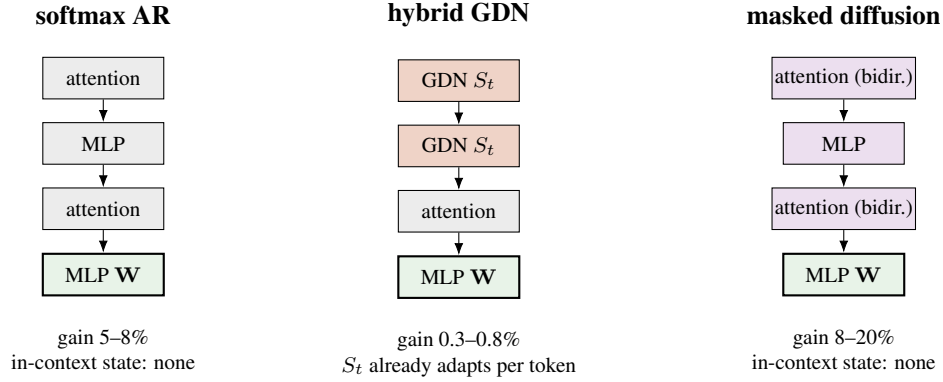

\textbf{Why the architecture decides.} A Gated-DeltaNet block keeps a fast-weight matrix, $S_t = S_{t-1}(I - \beta_t k_t k_t^\top) + \beta_t v_t k_t^\top$. The delta rule updates it as each token arrives. That is the same operation a gradient step on a linear layer performs, applied at every token with learned gates. \citet{ttt_linear_attention} show the equivalence formally. We measured a 0.3--0.8\% residual on hybrids. It is the part of document statistics the recurrent state does not already carry. Diffusion denoisers and autoregressive softmax models both lack such a state. Both gain. We drew the three cases side by side in Figure~\ref{fig:arch}.

\textbf{Why one objective destroys and the other preserves.} We ran next-token loss on a repeated sentence. It taught the output layer to emit that sentence. That collides with the output layer's role in every other generation. The collapse of the GSM8K answer marker is that collision made visible. A KL on a decision distribution teaches a decision. It leaves the generation pathway alone. The two writes differ in drift by less than a factor of two and in GSM8K cost by a factor of thirty.

\textbf{The design rule.} The measurements give a division of labour for a system that must keep learning after deployment. The gradient path belongs on softmax and diffusion backbones. It carries adaptation to the document at hand. That costs 1.3--1.5$\times$ a frozen pass, with per-document reset. On hybrid backbones the gradient path can be left off. The recurrent state already does this. Conclusions go into the weights as behaviour, through a gentle low-rank adapter trained on the system's own conclusion-conditioned judgments. The collateral probe gates every write. Facts go into an external store the model reads from context, because the one mechanism that stores them in these weights also breaks generation. Training the fast-weight pathway so that the frozen model learns to read what the gradient writes, and sourcing behaviour writes from inquiry on tasks where the model's own judgments beat its zero-shot answers, are the two directions this measurement sets up.

\section{Conclusion}

On the model families in use in 2026, we ran inference-time gradient adaptation. We used one untrained recipe and a protocol that separates context from weights. Softmax-attention and diffusion language models gained 5--20\% of their loss over a document. They outperformed full-context attention on the same text. Hybrid linear-attention models gained under 1\%. Their recurrent state already performs the adaptation. We checked retrieval at $n\ge100$ on two model sizes. The updated weights retrieved nothing on eight long-context and knowledge benchmarks. We wrote facts with the same objective. That stored format and broke reasoning at every dose. We wrote a conclusion as behaviour, sourced from an inquiry loop and distilled into a low-rank adapter. It reached the in-context ceiling with reasoning intact. For system designers, the adaptation finding shows where the gradient path pays and where the architecture has already paid for it. The write findings tell them what to put in weights and what to keep in context.

\bibliographystyle{plainnat}

\appendix

\section{Additional Results}
\label{app:more}

\begin{table}[htbp]
\caption{Learning-rate sweeps for every model on PG-19 (gain in nats per token, 95\% CI where $n \ge 10$, drift of $\fw$). Diffusion rows use the denoising loss and update clipping.}
\label{tab:lrsweep}
\centering\small
\begin{tabular}{llrrrr}
\toprule
Model & $n$ & $\eta$ & Gain & Improved & Drift \\
\midrule
Qwen3-1.7B & 20 & $3\times10^{-4}$ & $+0.121$ & 20/20 & 0.05\% \\
 & 20 & $10^{-3}$ & $+0.171$ & 20/20 & 0.05\% \\
 & 20 & $3\times10^{-3}$ & $+0.203$ & 20/20 & 0.12\% \\
 & 20 & $10^{-2}$ & $+0.219$ & 20/20 & 0.18\% \\
 & 20 & $3\times10^{-2}$ & $+0.209$ & 20/20 & 0.45\% \\
 & 50 & $10^{-2}$ & $+0.214 \pm 0.011$ & 50/50 & 0.19\% \\
 & 28 & $10^{-2}$ & $+0.216 \pm 0.011$ & 28/28 & 0.19\% \\
Qwen3-4B & 20 & $10^{-3}$ & $+0.156 \pm 0.015$ & 20/20 & 0.05\% \\
 & 20 & $10^{-2}$ & $+0.227 \pm 0.017$ & 20/20 & 0.20\% \\
Gemma-4-E4B & 20 & $3\times10^{-4}$ & $+0.143 \pm 0.019$ & 20/20 & 0.07\% \\
 & 20 & $10^{-3}$ & $+0.145 \pm 0.020$ & 20/20 & 0.17\% \\
Qwen3.5-4B & 20 & $3\times10^{-4}$ & $+0.006 \pm 0.001$ & 20/20 & 0.04\% \\
 & 20 & $10^{-3}$ & $+0.012 \pm 0.002$ & 20/20 & 0.12\% \\
 & 20 & $3\times10^{-3}$ & $+0.012 \pm 0.003$ & 20/20 & 0.32\% \\
 & 12 & $10^{-2}$ & $-0.02$ to $-0.08$ & 0/12 & 1.0\% \\
Qwen3.8-27B & 10 & $10^{-3}$ & $+0.008 \pm 0.002$ & 10/10 & \\
 & 10 & $10^{-2}$ & $+0.019 \pm 0.006$ & 10/10 & 0.21\% \\
Nemotron-Diffusion-8B & 1 & $10^{-2}$, no clip & diverged (9.3 $\rightarrow$ 41.6) & & 11\% \\
 & 10 & $10^{-3}$ & $+0.67$ & 8/10 & 0.3\% \\
 & 10 & $3\times10^{-4}$ & $+0.82$ & 10/10 & 0.12\% \\
DiffusionGemma-26B-A4B & 10 & $10^{-3}$ & $+2.42$ & 10/10 & 0.14\% \\
\bottomrule
\end{tabular}
\end{table}

\begin{table}[htbp]
\caption{Dose curve for storing one fact by the next-token objective (CounterFact, Qwen3-4B, 30 facts, recall on a rephrased question unless stated).}
\label{tab:dose}
\centering\small
\begin{tabular}{lrr}
\toprule
Write & Recall & Says old answer \\
\midrule
none & 0.03 & \\
fact in prompt & 0.83 & \\
1 step, $\eta=10^{-3}$ & 0.03 & \\
10 passes, $\eta=10^{-3}$ & 0.03 & \\
10 passes, $\eta=3\times10^{-3}$ & 0.10 & \\
50 passes, $\eta=10^{-3}$ & 0.23 & \\
20 passes, $\eta=10^{-2}$ & 0.50 & 0.07 \\
20 passes, $\eta=10^{-2}$, verbatim question & 1.00 & \\
20 passes, $\eta=10^{-2}$, 60\% of blocks writable & 0.37 & 0.07 \\
20 passes, $\eta=10^{-2}$, fact + 3 self-written QA pairs & 0.27 & 0.10 \\
\bottomrule
\end{tabular}
\end{table}

\begin{table}[htbp]
\caption{Interference across a session of 30 sequential fact writes (Qwen3-4B, 20 passes each at $\eta=10^{-2}$, no reset).}
\label{tab:session}
\centering\small
\begin{tabular}{lrrr}
\toprule
 & Rephrased recall & Verbatim recall & Says old answer \\
\midrule
before the session & 0.03 & 0.03 & 0.17 \\
after all 30 written, all 30 probed & 0.23 & 0.70 & 0.07 \\
first 10 written & 0.30 & 0.80 & \\
last 10 written & 0.30 & 0.90 & \\
one fact alone (Table~\ref{tab:dose}) & 0.50 & 1.00 & \\
\bottomrule
\end{tabular}
\end{table}

\begin{table}[htbp]
\caption{Behaviour writes on hybrid models (hidden-rule environment, 20 games each). All entries are AUROC; ``collateral'' is other-rule prompt-following before $\rightarrow$ after.}
\label{tab:hybridwrite}
\centering\small
\begin{tabular}{llrrrrr}
\toprule
Model & Write & Chance & Ceiling & Written & Collateral & Drift \\
\midrule
Qwen3.5-4B & late MLP, Adam $2\times10^{-5}$, oracle & 0.51 & 0.89 & 0.96 & $0.81\rightarrow0.67$ & 1.8\% \\
Qwen3.5-4B & late MLP, Adam $2\times10^{-5}$, teacher & 0.51 & 0.89 & 0.86 & $0.81\rightarrow0.72$ & 1.4\% \\
Qwen3.5-4B & late MLP, Adam $5\times10^{-6}$, teacher & 0.51 & 0.89 & 0.79 & $0.81\rightarrow0.80$ & 0.5\% \\
Qwen3.5-4B & LoRA r16, Adam $5\times10^{-5}$, teacher & 0.51 & 0.89 & 0.90 & $-0.04$ & \\
Qwen3.5-9B & late MLP, Adam $2\times10^{-5}$, teacher & 0.62 & 0.92 & 0.89 & $0.83\rightarrow0.80$ & 1.1\% \\
Qwen3.5-9B & LoRA r16, Adam $5\times10^{-5}$, teacher & 0.62 & 0.92 & 0.92 & $0.83\rightarrow0.83$ & \\
\bottomrule
\end{tabular}
\end{table}

\begin{table}[htbp]
\caption{Behaviour writes on real legal tasks (Qwen3-8B, raw option scoring). Teacher labels are the model's own few-shot judgments; oracle labels are the dataset labels.}
\label{tab:casehold}
\centering\small
\begin{tabular}{llrrrr}
\toprule
Task & Write & $n$ test & Zero-shot & Few-shot & After write \\
\midrule
CaseHOLD & late MLP, 256 teacher labels & 200 & 0.690 & 0.665 & 0.655 \\
CaseHOLD & late MLP, 256 oracle labels & 200 & 0.690 & & 0.670 \\
CaseHOLD & LoRA r16, 256 oracle labels & 200 & 0.690 & & 0.745 \\
CaseHOLD & LoRA r16, 1024 oracle labels, 3 epochs & 200 & 0.690 & & \best{0.775} \\
consumer\_contracts\_qa & late MLP, 196 teacher labels & 200 & 0.950 & 0.950 & 0.940 \\
\bottomrule
\end{tabular}
\end{table}

\begin{table}[htbp]
\caption{Judge quality for the inquiry loop: 50 games, 24 rules, true rule is the posterior mode (top-1), mean posterior on the true rule, and expected calibration error of $P(\text{yes})$.}
\label{tab:judge}
\centering\small
\begin{tabular}{lrrrr}
\toprule
Judge & Observations & Top-1 (oracle) & $p_{\text{true}}$ & ECE \\
\midrule
Qwen3-1.7B, zero-shot prompt & 8 & 0.06 (0.90) & 0.07 & 0.65 \\
Qwen3-1.7B, few-shot & 8 & 0.60 (0.90) & 0.61 & 0.18 \\
Qwen3-4B, few-shot & 8 & 0.68 (0.90) & 0.67 & 0.107 \\
Qwen3-8B, few-shot & 8 & 0.76 (0.90) & 0.70 & 0.098 \\
Qwen3-1.7B, few-shot & 16 & 0.64 (0.92) & 0.62 & 0.183 \\
Qwen3-4B, few-shot & 16 & 0.76 (0.92) & 0.77 & 0.108 \\
\bottomrule
\end{tabular}
\end{table}

\section{Claim--Evidence Map}
\label{app:claims}
\small
\begin{tabular}{p{0.46\linewidth}p{0.50\linewidth}}
\toprule
Claim & Evidence (run id in the released logs) \\
\midrule
Softmax models gain 5--8\% on every document and beat full 32k context & Table~\ref{tab:adapt}, Figures~\ref{fig:deciles}--\ref{fig:lr}; phase0-constant-set2/set3 (98 books), phase0-constant-qwen3-4b-lr1e-2, gemma4-e4b-lr1e-3, legal100 runs; control phase0-fullctx-20261009-190118 \\
Hybrid GDN models gain under 1\% at every rate & qwen35-4b at $3\times10^{-4}$, $10^{-3}$, $3\times10^{-3}$, $10^{-2}$; qwen38-27b at $10^{-3}$, $10^{-2}$ \\
Diffusion LMs gain 8--20\% with lowered rate and clipping & Table~\ref{tab:diffusion}, Figure~\ref{fig:diffusion}; phase0-diffusion-Nemotron-*-lr3e-4-clip, -lr1e-3-clip, -plainCE, -legal; diffusiongemma-*-canvas-* \\
Late blocks, constant rate, one step are each best & Table~\ref{tab:ablate}, Figure~\ref{fig:ablate}; phase0-constant-first-quarter, -middle-quarter, phase0-sigmoid\_z, phase0-clip, phase0-constant-k4 \\
Zero retrieval recovery at $n\ge100$ on two sizes & Table~\ref{tab:retrieval}, Figure~\ref{fig:retrieval}; rawchoice-27b, rawchoice-4b, n100b-27b, n100b-4b, babilong100-27b, lbv2-raw150-27b, cnli-27b, cnli-4b, citation-* \\
Fact writes store format and cost 36--38 GSM8K points & Table~\ref{tab:facts}, Figure~\ref{fig:facts}; cf-verbatim-lr1e-2-p20, zsre-27b, zsre-chat-27b, session-writes-* with retention/ \\
Behaviour write reaches the in-context ceiling, GSM8K intact & Table~\ref{tab:behaviour}, Figure~\ref{fig:dose}; distill-lora-teacher-20261010-095817 and -095818 (100 games), retention/gsm8k-behaviourwrite-8b \\
Text write of a conclusion is at chance & coupling-20261009-212352, -212522, -213159 \\
Inquiry loop: judge quality and 31\% fewer questions & Figure~\ref{fig:inquiry}, Table~\ref{tab:judge}; h1-20261009-185254 (8B), h4-20261009-182856 \\
CaseHOLD $+8.5$ with 1024 true labels through LoRA & Table~\ref{tab:casehold}; distill-task-casehold-lora-oracle-20261010-064720 \\
\bottomrule
\end{tabular}

\end{document}